\documentclass[lettersize,journal]{IEEEtran}
\usepackage{amsmath,amsfonts}
\usepackage{algorithmic}
\usepackage{algorithm}
\usepackage{array}
\usepackage[caption=false,font=normalsize,labelfont=sf,textfont=sf]{subfig}
\usepackage{textcomp}
\usepackage{stfloats}
\usepackage{url}
\usepackage{verbatim}
\usepackage{graphicx}
\usepackage{tabularx}
\usepackage{cite}
\usepackage[utf8]{inputenc} 
\usepackage[T1]{fontenc}    
\usepackage{hyperref}       
\usepackage{url}            
\usepackage{booktabs}       
\usepackage{amsfonts}       
\usepackage{nicefrac}       
\usepackage{microtype}      
\usepackage{xcolor}         
\usepackage{wrapfig}

\usepackage{multirow,multicol}
\usepackage{bbm}
\usepackage{bm}
\usepackage{float}
\usepackage{colortbl}
\usepackage{tabu}
\usepackage{amsmath}
\usepackage{amssymb}
\usepackage{bm}
\usepackage{graphicx}

\begin{document}

\title{PhyTracker: An Online Tracker for Phytoplankton}

\author{Yang Yu, Qingxuan Lv, Yuezun Li,~\IEEEmembership{Member,~IEEE,} Zhiqiang Wei, Junyu Dong,~\IEEEmembership{Member,~IEEE,}
\thanks{Yuezun Li and Junyu Dong are \textit{corresponding authors}.}
\thanks{Yang Yu, Qingxuan Lv, Yuezun Li, Zhiqiang Wei, and Junyu Dong are with the College of Computer Science and Technology, Ocean University of China, China. e-mail: (oucyuyang@stu.ouc.edu.cn; lvqingxuan@stu.ouc.edu.cn; liyuezun@ouc.edu.cn; weizhiqiang@ouc.edu.cn; dongjunyu@ouc.edu.cn).
Copyright © 2024 IEEE. Personal use of this material is permitted. However, permission to use this material for any other purposes must be obtained from the IEEE by sending an email to pubs-permissions@ieee.org.}
}

\markboth{ IEEE TRANSACTIONS ON CIRCUITS AND SYSTEMS FOR VIDEO TECHNOLOGY}%
{Shell \MakeLowercase{\textit{et al.}}: A Sample Article Using IEEEtran.cls for IEEE Journals}


\maketitle

\begin{abstract}
Phytoplankton, a crucial component of aquatic ecosystems, requires efficient monitoring to understand marine ecological processes and environmental conditions. Traditional phytoplankton monitoring methods, relying on non-in situ observations, are time-consuming and resource-intensive, limiting timely analysis. To address these limitations, we introduce PhyTracker, an intelligent in situ tracking framework designed for automatic tracking of phytoplankton. PhyTracker overcomes significant challenges unique to phytoplankton monitoring, such as constrained mobility within water flow, inconspicuous appearance, and the presence of impurities. Our method incorporates three innovative modules: a Texture-enhanced Feature Extraction (TFE) module, an Attention-enhanced Temporal Association (ATA) module, and a Flow-agnostic Movement Refinement (FMR) module. These modules enhance feature capture, differentiate between phytoplankton and impurities, and refine movement characteristics, respectively. Extensive experiments on the PMOT dataset validate the superiority of PhyTracker in phytoplankton tracking, and additional tests on the MOT dataset demonstrate its general applicability, outperforming conventional tracking methods. This work highlights key differences between phytoplankton and traditional objects, offering an effective solution for phytoplankton monitoring.
\end{abstract}


\section{Introduction}

\begin{figure}[!t]
  \centering
  \scalebox{1.0}{
  \includegraphics[width=0.9\linewidth]{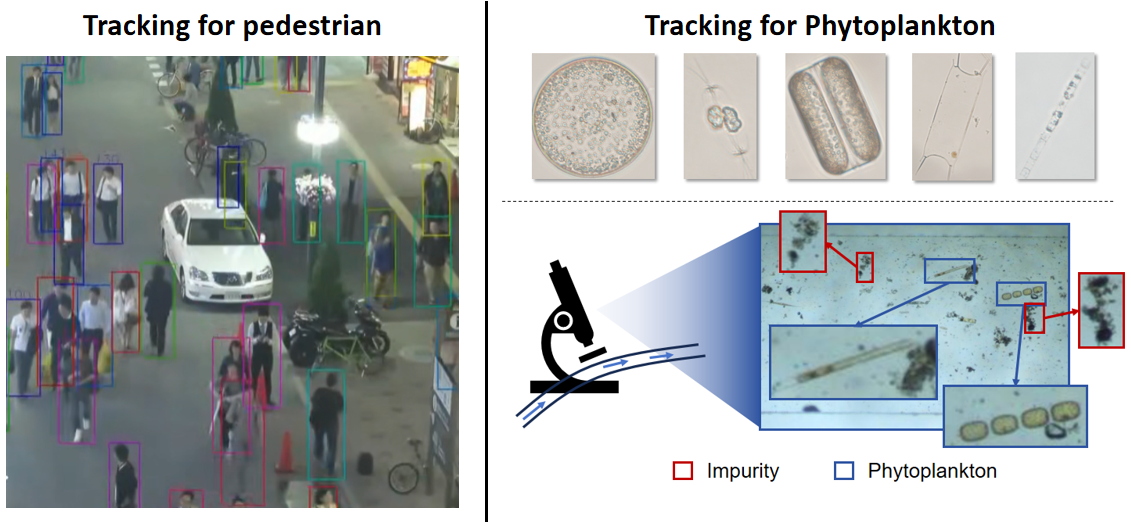}}
    \vspace{-0.1cm}
  \caption{\small The pedestrian dataset (on the left) and the planktonic dataset (on the right) have different characteristics.}
  \label{fig:example}
\end{figure}

\IEEEPARstart{P}{hytoplankton} is a general term for plant microorganisms, particularly referring to microalgae (see Fig.~\ref{fig:example}) ~\cite{matsunaga2005marine}. Phytoplankton is a vital component of aquatic ecosystems, with their activities serving as key indicators for marine ecological processes and environmental conditions~\cite{chen2018microalgae}. Consequently, monitoring phytoplankton holds significant importance in maintaining the stability of aquatic ecosystems, safeguarding water resources, and advancing scientific exploration~\cite{ahmad2021role}. 

Traditional efforts on monitoring phytoplankton mainly rely on the so-called \textit{non-in situ observation} approach, that is to collect water samples and bring them back to the laboratory with manual observation~\cite{alvarez2011effectively}. This approach not only consumes considerable time and human resources but also fails to analyze phytoplankton timely. To overcome this limitation, we develop an intelligent tracking framework, called \textit{PyTracker}, that can be deployed on the ocean to monitor phytoplankton in a way of \textit{in situ observations}. This framework is designed to automatically localize and categorize phytoplankton and then track them constantly observed in the microscope. The results can provide versatile information in monitoring phytoplankton, and can be utilized for further analysis such as density estimation, action recognition, pose estimation, etc. 

However, tracking phytoplankton is significantly more challenging compared to the scenario of tracking general objects on the ground, which mainly lies in the following three aspects:
\begin{enumerate}
    \item \textbf{Inconspicuous appearance:} Phytoplankton commonly exhibit tiny sizes, light colors, erratic forms, and straightforward textures. These characteristics significantly increase the challenge of identification than general objects on the ground. 

    \item \textbf{Complex monitor scenario:} The water samples usually contain impurities dispersed throughout the entire view of observation. These impurities look very similar to phytoplankton, posing challenges to the accurate monitoring of phytoplankton. 

    \item \textbf{Different monitor pipeline:} 
    Tracking objects on the ground, particularly pedestrians and vehicles, typically utilizes general video cameras and is performed under a wild scenario. However, the pipeline of our task is notably different, which involves extracting water samples from the ocean and gradually passing them through the microscopes for analysis. Under this scenario, the mobility of phytoplankton is highly constrained, and their movement within this pipeline is mainly driven by water flow, resulting in a highly uniform trajectory of these phytoplankton. 
\end{enumerate}

These differences greatly limit the application of conventional tracking methods for ground monitoring to phytoplankton monitoring~\cite{zhang2015tracking,zhang2021finding,smith2010cooperative}. 
As such, developing a tracking framework devoted to this task is necessary.  

In this paper, we conduct an in-depth analysis of the unique characteristics of phytoplankton and proposes a new method called \textit{PhyTracker} devoted to accomplishing the tracking of phytoplankton. Our method achieves tracking in an online manner, which continuously track the phytoplankton alongside water flows with a meticulously designed architecture. Specifically, our method features three designs: 
\textbf{1)} Since the appearance of phytoplankton is inconspicuous, we describe a Texture-enhanced Feature Extraction (TFE) module to improve the capture of appearance features, with the incorporation of dilated convolutions and SRM filters.
\textbf{2)} Floating impurities likely disrupt the temporal association of phytoplankton, causing the chaotic target correlation between frames before and after the tracking process. To mitigate this, we propose an Attention-enhanced Temporal Association (ATA) module to tell apart phytoplankton and impurities. The core of this module is an attention mechanism, which can effectively associate corresponding features from consecutive frames, eliminating the interference caused by impurities.
\textbf{3)} Trajectories are an important characteristic of phytoplankton. However, in our monitor pipeline, trajectories of phytoplankton are highly consistent, concealing the characteristic of phytoplankton movement. Therefore, we describe a Flow-agnostic Movement Refinement (FMR) module, which can recover the characteristics of each phytoplankton movement, making phytoplankton more discriminative.

Extensive experiments are conducted on a large-scale public phytoplankton tracking dataset (PMOT)~\cite{yu2023pmot2023}, demonstrating the superiority of our method in tracking phytoplankton. Moreover, we validate our method on the general object tracking dataset (MOT)~\cite{milan2016mot16} in comparison to recent conventional tracking methods. The results surprisingly corroborate that our method can still outperform others.

The contributions of this paper can be summarized in three-fold: 
\begin{enumerate}
    \item We thoroughly examine the major differences between traditional objects (such as pedestrians and vehicles) and phytoplankton, highlighting three key aspects: different monitor pipelines, inconspicuous appearance, and complex monitor scenarios. 

    \item To address these challenges, we propose an online tracker with three key improvements: a Texture-enhanced Feature Extraction (TFE) module, an Attention-enhanced Temporal Association (ATA) module and a Flow-agnostic Movement Refinement (FMR) module. Each module is developed to handle corresponding challenge. 

    \item We conduct comprehensive experiments on both the phytoplankton dataset and the general object tracking dataset. The results demonstrate the effectiveness of the proposed method, particularly in the context of phytoplankton tracking.
    
\end{enumerate}

\begin{figure*}[!t]
  \centering
  \includegraphics[width=\linewidth]{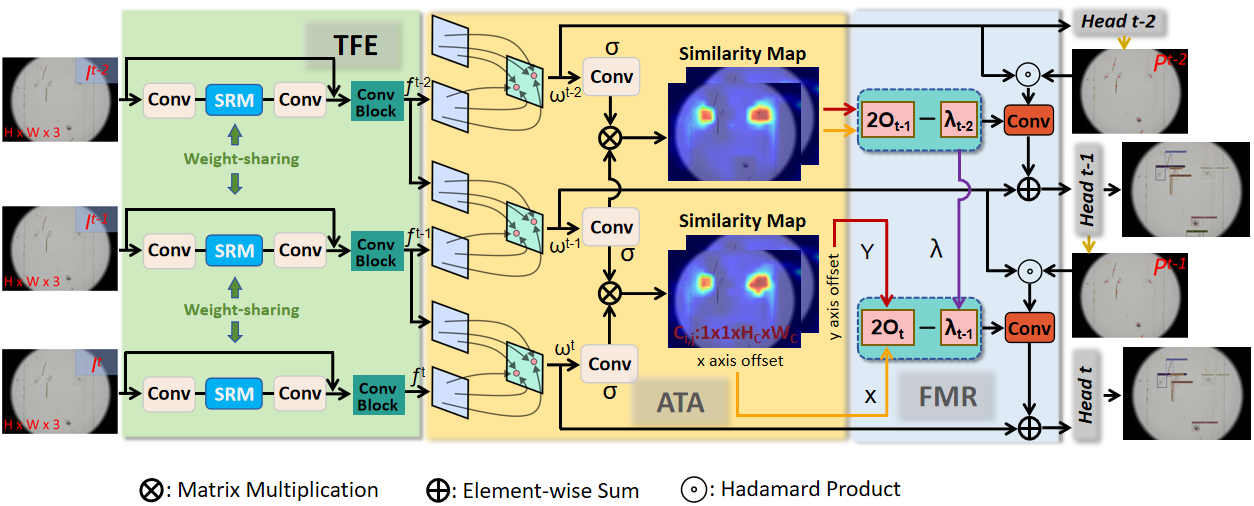}
  \vspace{-0.4cm}
  \caption{\small Overview of \textit{PhyTracker}. \textit{PhyTracker} receives the current frame picture and the previous frame picture at a time, as well as the trajectory information integrated from all previous frames, to assist in the tracking of the current frame. 
  The yellow arrows indicate the offset in the x-coordinate calculated from the Similarity Map, while the red arrows represent the offset in the y-coordinate derived from the same map. The purple arrow represents the Memory Offset obtained from the previous frame when calculating the current frame. The golden arrow indicates the detection result generated from the previous frame's head, which contains only positional information without any class information.}
  \label{fig:overview}
\end{figure*}

\section{Backgrounds and Related Works}

\smallskip
\noindent\textbf{Phytoplankton.} 
Phytoplankton are plant-like life forms (see Fig.\ref{fig:example}) that play an indispensable role in marine ecosystems. Through photosynthesis, they absorb carbon dioxide and release oxygen, providing a crucial source of oxygen for marine life, while also fixing carbon in organic matter~\cite{anderson1995hydrogen}. When they die, some of the organic carbon settles to the ocean floor, forming sediments and participating in long-term carbon storage and the Earth's carbon cycle~\cite{sellner1987phytoplankton}.
Phytoplankton are also the foundation of the marine food chain. They are consumed by zooplankton and other organisms, thereby supporting the entire marine ecosystem~\cite{fenchel1988marine}.
Additionally, phytoplankton play a role in regulating the global climate by modulating the reflectivity of the ocean surface, and by absorbing and releasing heat. They also impact the chemical composition of the atmosphere, thus affecting atmospheric circulation and climate. 

\smallskip
\noindent\textbf{Monitoring Phytoplankton.} 
Observing and real-time monitoring of phytoplankton species, density, and concentration have significant implications for humans and nature~\cite{boyce2015patterns,reynolds1984phytoplankton,irwin2015phytoplankton}.
Firstly, changes in phytoplankton species and density can reflect the ecological health of water bodies. By monitoring phytoplankton, we can promptly detect abnormal changes in ecosystems and take appropriate measures~\cite{tett2008use}.
Secondly, phytoplankton are very sensitive to environmental factors such as light, temperature, and carbon dioxide. Monitoring changes in phytoplankton can provide valuable data on climate change~\cite{winder2012phytoplankton}
Moreover, phytoplankton form the base of the marine food chain. Their density and distribution directly affect the reproduction and survival of other marine organisms. By monitoring phytoplankton, fishery managers can predict the density and distribution of fish resources, thereby formulating more effective fishery management strategies~\cite{boyce2015patterns}.
Additionally, certain species of phytoplankton can proliferate under specific conditions, forming harmful algal blooms (such as red tides), which lead to oxygen depletion in water bodies, release toxins, and pose threats to aquatic life and human health. Real-time monitoring of phytoplankton can provide early warnings of harmful algal blooms, reducing their negative impacts~\cite{smayda1997harmful}.

Traditional monitoring methods, referred to as \textit{non-in-situ observation}, involve the collection of water samples and their observation under microscopes by trained personnel~\cite{alvarez2011effectively,ciesielska2018observation,mccall1984systematic}. However, these techniques require a lot of time and human resources and lack the ability for timely phytoplankton analysis. 
Recently, advancements in hardware have made \textit{in-situ observation} feasible by integrating digital microscopes, flow pumps, and computational chips into a single device~\cite{joint2000estimation}. While this device shows promising potential for phytoplankton monitoring, the algorithms specifically dedicated to this task remains unexplored. Typically, tracking is the prerequisite task for monitoring phytoplankton. Given the tracking results, we can futher analyze the activities of phytoplankton. However, existing tracking algorithms are designed for ground scenario. 
Compared to the ground scenario, tracking phytoplankton poses unique challenges due to the different monitor pipelines, the inconspicuous appearance of phytoplankton, and the complexity of monitoring scenario. Therefore, there is an urgent need to develop a devoted tracking framework for phytoplankton.

\smallskip
\noindent\textbf{Object Tracking.}
The existing tracking methods are mainly designed for ground scenarios, focusing on tracking general objects such as vehicles and pedestrians~\cite{9142255,sun2016visual,zhu2021rgbt,yao2018semantics,li2018learning}. To obtain the trajectory of objects, Object extraction, temporal association and motion prediction are three important aspects in multi-object tracking within video sequences. According to the tracking pipeline, existing methods can be divided into two categories: Offline tracking and Online tracking.

\textit{Offline Tracking.} 
Offline tracking allows the use of information from subsequent frames and is formulated as a graph model for a globally optimal solution~\cite{jordan2004graphical,1708000,7041176}. 
However, this setup makes it not suitable for practical applications due to its reliance on future frame data.

\textit{Online Tracking.} 
Online multi-object tracking involves calculating the match between current object detection and existing trajectories based on the available data~\cite{9127935,8853333,8540936,10468656}. The nature of online tracking requires that the decision for each frame's tracking outcome must rely solely on information from the current and previous frames. It is imperative that the algorithm cannot use the current frame's data to alter the results of previous frames. Therefore, this paper focuses on the setting of onlione tracking.

\begin{figure}[!t]
  \centering
  \includegraphics[width=0.85\linewidth]{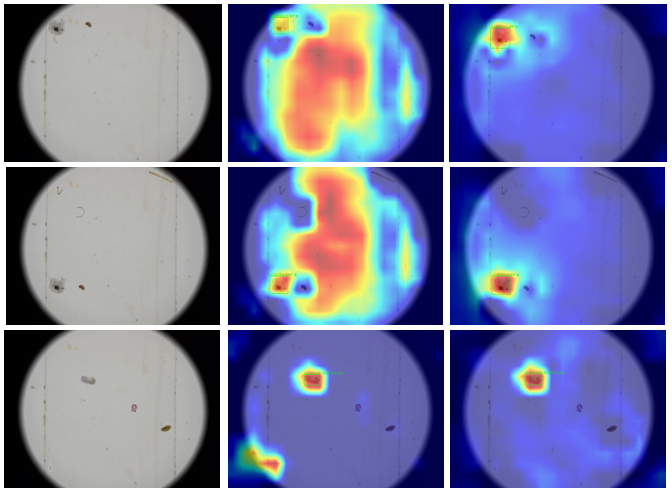}
  \vspace{-0.2cm}
  \caption{\small Each row displays the focus points of the algorithm on a single target only. The first column is the original image, the second column is DLA34+ByteTrack, and the third column is \textit{PhyTracker}.}
  \label{fig:heatmap}
\end{figure}

\section{Motivation and Preliminary Validation}
The straightforward solution for tracking phytoplankton is to directly adapt existing tracking methods into this task.
However, these methods are designed for ground scenarios and focus on tackling common challenges such as occlusions, target matching between frames, and camera jitters. Thus they do not align with the scenario of phytoplankton in aquatic environments. 
Unlike conventional tracking targets such as pedestrian and vehicles, phytoplankton exhibits significant differences in data characteristics, including their unique low-contrast color features, the stark contrast between aquatic and ground environments, and their distinct movement patterns compared to other organisms. These differences significantly hinder the application of existing methods to phytoplankton monitoring.
To validate this, we adopt the recent tracking methods ByteTrack~\cite{zhang2022bytetrack} to our task. 
As shown in Fig.~\ref{fig:heatmap}, it can be seen that the generated attention heatmaps of ByteTrack can hardly concentrate on the phytoplankton, indicating the infeasibility of directly adopting existing tracking methods.

\section{Method}
This paper describes an online tracker, \textit{PhyTracker}, devoted to monitoring phytoplankton. Compared to existing tracking methods, our method features three major improvements. First, we introduce a Texture-enhanced Feature Extraction (TFE) module to enhance the appearance distinction of phytoplankton. This improvement enables the phytoplankton becoming more detectable. Second, we propose an Attention-enhanced Temporal Association (ATA) module, which optimizes the feature distances between targets in adjacent frames, enhancing the capacity of model to distinguish between similar phytoplankton, as well as impurities and phytoplankton. Furthermore, we introduce a Flow-agnostic Movement Refinement (FMR) module, which effectively reduces feature confusion from similar motion trajectories between different tracking entities and preserves original movement offset information, thereby enhancing sensitivity to individual movement characteristics. These three improvements correspond to solving the three difficulties in phytoplankton monitoring, as described in Sec.\ref{fig:example}.

\subsection{Problem Setup}
Denote a video sequence with total $N$ frames as $\mathcal{V} = \{\mathcal{I}_t\}_{t=1}^N$, where $\mathcal{I}_t \in \mathbb{R}^{h \times w \times 3}$ refers to the $t$-th frame. Suppose this video sequence contains $\mathcal{D}$ phytoplankton. 
The goal of our method is to output the trajectories of all phytoplankton $\{ \mathcal{T}_i \}_{i=1}^{\mathcal{D}}$ within the video sequence $\mathcal{V}$. The $i$-th trajectory is a collection of bounding boxes at corresponding frames and its class label, defined as $\mathcal{T}_i = \{(b_{i,t_1},...,b_{i,t_N}),y_i \}$, where $b_{i,t_j}$ is the bounding box of $i$-th phytoplankton at temporal index $t_j$, $t_N$ is the length of trajectory and $y_i$ represents category.

\subsection{Framework Workflow}
The workflow of our method is inspired by TraDes~\cite{trades}. 
Given a frame $\mathcal{I}_t$, it is first passed into Texture-enhanced Feature Extraction (TFE) module to extract appearance features of phytoplankton as $f^t \in \mathbb{R}^{\frac{h}{4} \times \frac{w}{4} \times 64}$. Then the feature of current frame $f^t$ and the one of previous frame $f^{t-1}$ are sent into the Attention-enhanced Temporal Association (ATA) module, to build the correlations between these two features. Based on the correlations, ATA can predict the movement offsets of phytoplankton. Then the movement offsets are forwarded into Flow-agnostic Movement Refinement (FMR) module, which eliminates the effect of water flows and fuses the knowledge from previous frames into the head network for final prediction.

\begin{figure}[!t]
  \centering
  \includegraphics[width=\linewidth]{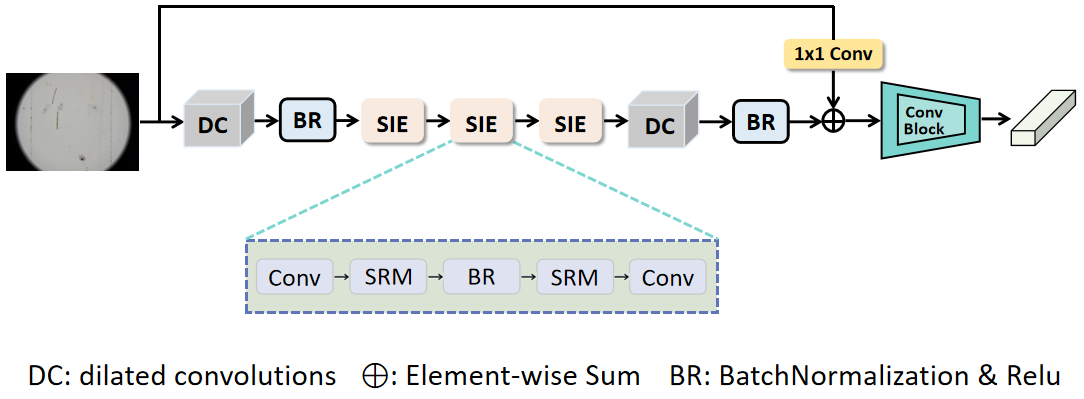}
  \vspace{-0.3cm}
  \caption{\small We use three consecutive SIE modules(Semantic Information Extraction module) to extract semantic feature. The core of the SIE network is the SRM filter.}
  \label{fig:module1}
\end{figure}

\subsection{Texture-enhanced Feature Extraction}
Phytoplankton often exhibit appearance similar to their natural aquatic environments, making it difficult for traditional methods to capture discriminative features.
In light of this, we propose a Texture-enhanced Feature Extraction (TFE) module to enhance feature extraction of phytoplankton (see Fig.~\ref{fig:module1}). Inspired by~\cite{quan2023centralized,hashmi2023featenhancer}, this module extracts the noise information combined with semantic information to enrich the representation of phytoplankton features. Specifically, to amplify the subtle textures, we employ dilated convolutions~\cite{yu2017dilated} to expand the receptive field without increasing computational load. Then several SIE blocks are proposed to refine the features. SIE block is composed of convolution layers and SRM layers. SRM filters were originally designed to address the issues of image denoising and edge preservation in image processing. Under the microscope, there are situations where the flow image is unclear and there are many impurities. In response to this, we have modified the SRM filter to match the effectiveness of extracting additional information from phytoplankton data. The SRM layer is three 5$\times$5 convolutional kernels with fixed values that remain unchanged, the kernels are:
\begin{equation}
\footnotesize 
{
\label{eq: srm}
\setlength{\arraycolsep}{1.2pt}
\begin{aligned}
\small
&\begin{bmatrix}
    0 & 0 & 0 & 0 & 0 \\
    0 & -1 & 2 & -1 & 0 \\
    0 & 2 & -4 & 2 & 0 \\
    0 & -1 & 2 & -1 & 0 \\
    0 & 0 & 0 & 0 & 0 
\end{bmatrix}
\begin{bmatrix}
    -1 & 2 & -2 & 2 & -1 \\
    2 & -6 & 8 & -6 & 2 \\
    -2 & 8 & -12 & 8 & -2 \\
    2 & -6 & 8 & -6 & 2 \\
    -1 & 2 & -2 & 2 & -1 
\end{bmatrix}
\begin{bmatrix}
    -1 & -1 & -1 & -1 & -1 \\
    -1 & 0 & 0 & 0 & -1 \\
    -1 & 0 & 8 & 0 & -1 \\
    -1 & 0 & 0 & 0 & -1 \\
    -1 & -1 & -1 & -1 & -1 
\end{bmatrix}
\end{aligned}
}
\end{equation}


The features from SIE blocks are then integrated with the original images, which are sent into a DLA34 network~\cite{yu2018deep}. This module can reveal differences in textures between phytoplankton and the aquatic environment, which are not easily observable in the RGB space.

\begin{figure*}[!t]
  \centering
  \includegraphics[width=0.9\linewidth]{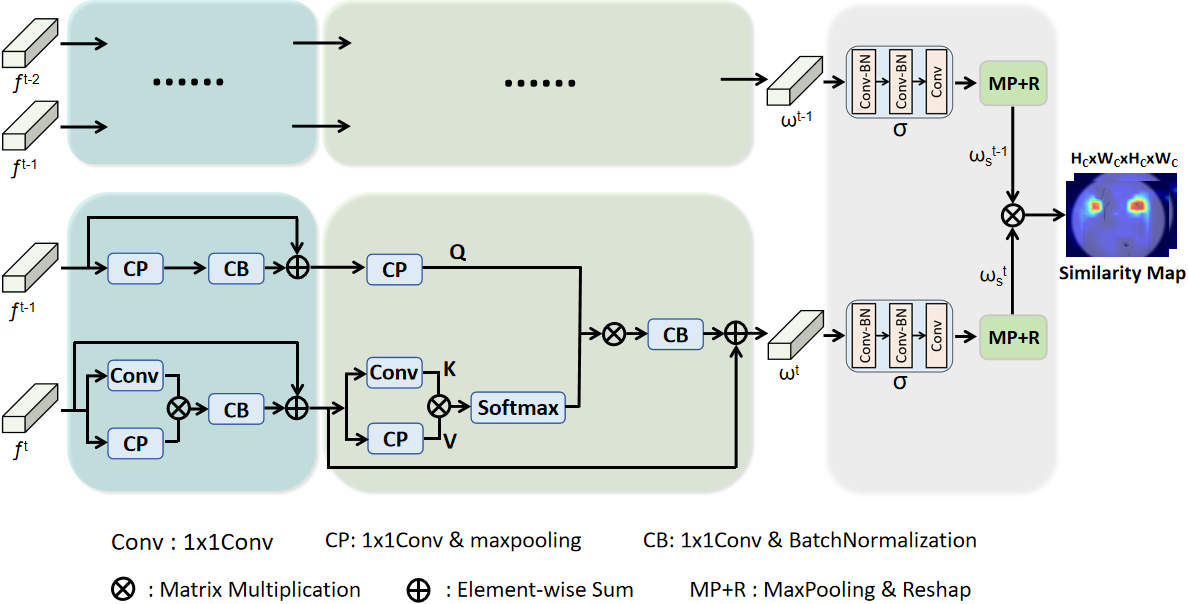}
  \vspace{-0.3cm}
  \caption{\small We take the basic features of three consecutive frames as input and output the similarity information between frames.}
  \label{fig:module2}
\end{figure*}

\subsection{Attention-enhanced Temporal Association}

\begin{figure}[!t]
  \centering
  \includegraphics[width=0.9\linewidth]{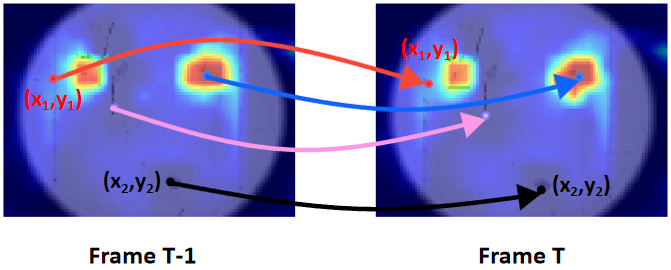}
  \vspace{-0.3cm}
  \caption{\small Each line in the graph represents the similarity between points.}
  \label{fig:similarity}
\end{figure}

\begin{figure}[!t]
  \centering
  \includegraphics[width=0.9\linewidth]{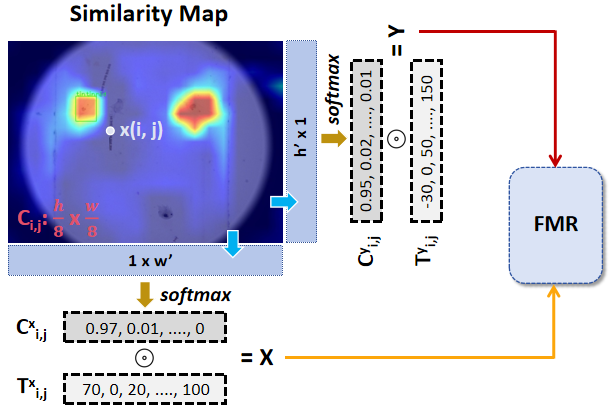}
  \vspace{-0.3cm}
  \caption{\small The figure shows the calculation process of generating offsets from the Similarity Map, and the generated offsets are stored in the $\mathcal{O}_t$ in FMR.}
  \label{fig:module2_last}
\end{figure}

Effectively associating the features of phytoplankton across frames is crucial for accurate tracking results. 
However, temporal association in our task is challenging, due to the widespread impurities in observing scenario and similar-appearance phytoplankton of different types.

To address this issue, we propose an Attention-enhanced Temporal Association (ATA) module, which is designed to effectively find the feature association of same target across consecutive frames.
The core of this module is a newly proposed Two-Stage Cross-Attention (TSCA) operations based on existing self-attention mechanism~\cite{vaswani2017attention,shen2021efficient}. As shown in Fig.~\ref{fig:module2}, the first stage is a feature refinement block, which enhances the features of current frame $f^t$ and previous frame $f^{t-1}$. Then the enhanced features are executed a cross-attention operation to model the temporal associations.  

Specifically, in the first stage, the feature $f^{t-1}$ is sent into the CP and CB modules, and then added with $f^{t-1}$ for refinement. Different from $f^{t-1}$, the feature $f^{t}$ is separately processed by Conv and CP modules, and calculates attention by matrix multiplication. 
The multiplication result is then sent into CB block, and perform residual connection with $f^t$ for refinement. In the second stage, the refined features are performed similar operations. But the difference is that we performance cross-attention operations to obtain the intermediate feature $\omega^t$. 
Let $Q$ represent query features obtained from $f^{t-1}$, and $K,V$ represent key and value features obtained from $f^{t}$. Inspired by~\cite{shen2021efficient}, the cross-attention operation can be defined as

\begin{equation}
  \textrm{CA}(Q, K, V) = {\phi_q}(Q)({\phi_k}(K)^TV),
\end{equation}
where $\phi_q$ and $\phi_k$ are the normalization functions for query and key features, implementing by normalization methods:

\begin{equation}
  \begin{aligned}
        & {\phi_q}(Q) = {\textrm{softmax}_{row}}(Q)  \\
        & {\phi_k}(K) = {\textrm{softmax}_{col}}(K),
       \end{aligned}
\end{equation}
note that $\textrm{softmax}_{row}$, $\textrm{softmax}_{col}$ denotes the application of the softmax function along each row or column of the corresponding input.

It is important to note that the feature $\omega^t$ represents a feature that enhances the association of the same tracking target in the current frame feature $f^t$ and the past frame feature $f^{t-1}$.

Inspired by~\cite{trades}, we then seek the similarity between adjacent features $\omega^{t-1}$ and $\omega^t$ to obtain higher level associations. These two features are sent into a convolution block $\sigma$ and then perform matrix multiplication with each other to create a four-dimensional similarity matrix $\mathcal{S} \in \mathbb{R}^{h' \times w' \times h' \times w'}$, where $h'=\frac{h}{8}, w'=\frac{w}{8}$. The element $\mathcal{S}(i,j,k,l)$ denotes the similarity between the location of $(i,j)$ in the current frame $t$ and the location of $(k,l)$ in the previous frame $t-1$ (see Fig.~\ref{fig:similarity}).

Based on this similarity matrix $\mathcal{S}$, we can generate the offset information  following~\cite{trades}. As shown in Fig.~\ref{fig:module2_last},
for a phytoplankton centered at the location of $(i, j)$ in the current frame $t$, we can fetch its two-dimensional similarity map $\mathcal{C}_{i,j} \in \mathbb{R}^{h' \times w'}$ from matrix $\mathcal{S}$. This similarity map $\mathcal{C}_{i,j}$ stores the similarities among the phytoplankton and all locations in the previous frame $t-1$. To calculate the offset, $\mathcal{C}_{i,j}$ is first max pooled by $1 \times w'$ and $h' \times 1$ kernels and then normalized by a softmax function to obtain two vectors, $\mathcal{C}^{\mathcal{X}}_{i,j} \in [0,1]^{1 \times w'}$ and $\mathcal{C}^{\mathcal{Y}}_{i,j} \in [0,1]^{h' \times 1}$, respectively. These two vectors represent the likelihood of this phytoplankton appearing on horizontal and vertical locations in frame $t$. Then we create two offset templates $\mathcal{T}^{\mathcal{X}}_{i,j} \in \mathbb{R}^{1 \times w'}, \mathcal{T}^{\mathcal{Y}}_{i,j} \in \mathbb{R}^{h' \times 1}$ in the horizontal and vertical directions, respectively, which are calculated by 
\begin{equation}
  \begin{aligned}
        \mathcal{T}^{\mathcal{X}}_{i,j}(l) & = (l - j) \times s, & 1 \leq l \leq w' \\
        \mathcal{T}^{\mathcal{Y}}_{i,j}(k) & = (k - i) \times s, & 1 \leq k \leq h'
       \end{aligned}
\end{equation}
where $s$ is the feature stride of $\omega_s^t$. $\mathcal{T}^{\mathcal{X}}_{i,j}(l)$ refers to the horizontal offset when the phytoplankton appears at the location of $(:,l)$ in frame $t-1$. 
Let $\mathcal{O}_t = [ \mathcal{O}^{\mathcal{X}}_t, \mathcal{O}^{\mathcal{Y}}_t]$ be the offset information, containing the horizontal and vertical offsets respectively. Each offset can be inferred by the dot product between the likelihoods and actual offset values as
\begin{equation}
    \mathcal{O}^{\mathcal{X}}_t = \mathcal{C}_{i,j}^{\mathcal{X}} \mathcal{T}^{\mathcal{X}}_{i,j}, \;
    \mathcal{O}^{\mathcal{Y}}_t = \mathcal{C}_{i,j}^{\mathcal{Y}} \mathcal{T}^{\mathcal{Y}}_{i,j}.
\end{equation}

By applying this offset information, we can obtain the temporal association between the previous frame and the current frame. This offset information is then sent into the third module FMR to eliminate the impact of displacement caused by phytoplankton in the past frames.

\begin{figure}[!t]
  \centering
  \includegraphics[width=\linewidth]{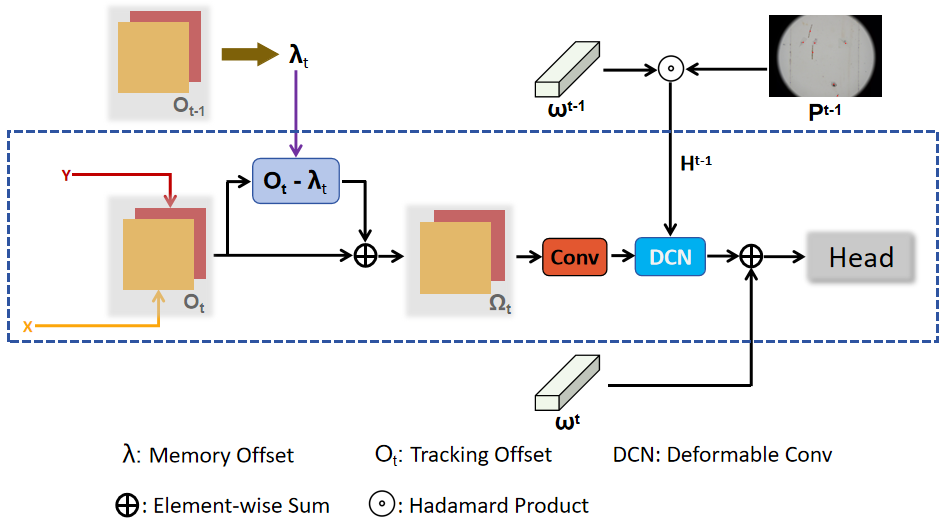}
  \vspace{-0.6cm}
  \caption{\small The interior of the dashed line represents the FMR module at frame t, while the rest represents external variables.}
  \label{fig:module3}
\end{figure}

\subsection{Flow-agnostic Movement Refinement}
The motion characteristics of phytoplankton under this scenario are mainly driven by water currents, exhibiting highly consistent movement trajectory characteristics within the pipeline. This phenomenon may lead to an undue reduction of feature differences of different phytoplankton classes, degrading the tracking performance. 

To this end, we propose a Flow-agnostic Movement Refinement (FMR) module (see Fig.~\ref{fig:module3}). This module aims to eliminate the effect of similar motion trajectories caused by water flow and amplify the differentiation between different phytoplankton individuals at the feature level. 
To achieve this, we need to first estimate the movement of water flow. Since the movement of all phytoplankton is driven by water flow, we can average the movement of all phytoplankton over all frames to represent water movement. Nevertheless, our framework is running online, which means it is impossible to access the future frames. To compensate, we maintain an offset memory bank and store the offset $\mathcal{O}_t$ constantly. Then we can average the offset in the bank as $\lambda_t = \frac{1}{t} (\mathcal{O}_t + ... + \mathcal{O}_0)$. In the current frame $t$, we subtract $\lambda$ from the offset $\mathcal{O}_t$ to obtain the movement characteristics of phytoplankton, which serves as an extra feature added with the original offset $\mathcal{O}_t$ to form flow-agnostic feature $\Omega_t$. This process can be written as $\Omega_t = 2\mathcal{O}_t - \lambda_t$. 

Then we propagate the features $\mathcal{H}^{t-1}$ from the previous frame to the current refined offset $\Omega_t$ using a deformable convolution~\cite{dai2017deformable}, inspired by~\cite{trades}. The feature $\mathcal{H}^{t-1}$ is calculated as
\begin{equation}
    \mathcal{H}^{t-1} = \omega^{t-1} \circ \mathcal{P}^{t-1},
\end{equation}
where $\circ$ is the Hadamard product, $\mathcal{P}^{t-1}$ is the class-agnostic center heatmap produced from the head network (we use the same head network as CenterNet~\cite{zhou2019objects}).

\section{Experiments}

\subsection{Experimental Settings}

\begin{figure*}[!t]
  \centering
  \includegraphics[width=1.0\linewidth]{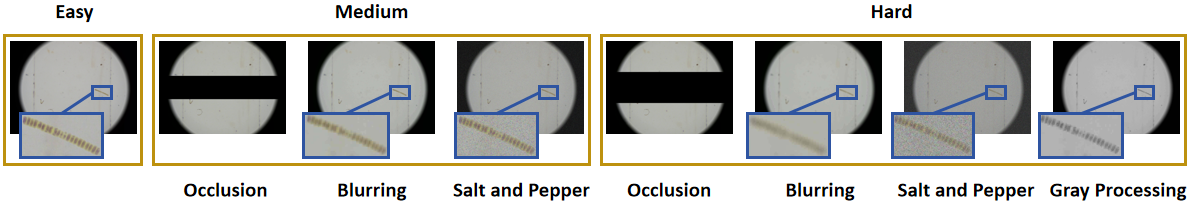}
  \caption{\small The three rows represent easy, medium, and hard difficulty levels, with no noise added in the easy level. Each column represents a type of added noise, with no gray processing present in moderate difficulty.}
  \label{fig:compare3}
\end{figure*}

\noindent\textbf{Datasets.}
Our method is validated on two public datasets, PMOT~\cite{yu2023pmot2023} and MOT~\cite{milan2016mot16}. 
\begin{enumerate}
    \item The PMOT dataset is a synthetic dataset specifically designed for phytoplankton tracking tasks, encompassing a total of 21 categories. It simulates video footage of plankton observed in flowing pipes under a microscope, making it the first synthetic dataset of its kind. Tracking phytoplankton in real-world environments involves much more complex scenes compared to laboratory settings. Therefore, we expanded our dataset to simulate the presence of noise found in real-world scenarios. 
    The phytoplankton dataset we use is derived from modifications to the PMOT2023 dataset. We integrated data collected over the years from our laboratory, selecting the most suitable portions for inclusion. Based on this, we applied transformations such as occlusions, gray processing, blurring and salt-and-pepper noise to simulate complex underwater environmental conditions. Depending on the degree of added noise, we classified the entire phytoplankton dataset into three difficulty levels: no noise for easy difficulty; a lower degree of noise for medium difficulty; and a higher degree of noise for hard difficulty. As shown in Fig.~\ref{fig:compare3}. This allowed us to validate the model's generalization capabilities under more realistic conditions. 
    The entire dataset consists of 9 original video segments and 63 noise-added video segments, with the length of the transformed videos matching that of the original videos. The composition of the dataset is shown in Table~\ref{table_dataset}. The ablation studies were evaluated using the phytoplankton dataset. 
    \item To fully demonstrate the effectiveness of our method, we conducted experiments on the MOT dataset, a widely recognized benchmark in the field of multi-object tracking. The MOT dataset is designed to present a range of complex scenarios, including busy streets, shopping malls, parks, and other environments, and is provided by the MOT Challenge organization.  
    The MOT17 dataset includes annotations for the training set but does not provide annotations for the test set; performance metrics can only be obtained by uploading the tracking results to the MOTChallenge website for evaluation. In our experiments with the MOT17 dataset, we performed an overall algorithm evaluation using only the training set. Specifically, we divided the training set into two halves: one half was used for training, and the other half was used for testing. 
\end{enumerate}

\begin{table*}[!t]
\centering
\renewcommand{\arraystretch}{1.4}
\caption{\small During training, we only use the original images from the training set for training.}
\scalebox{1.0}{
\resizebox{\textwidth}{!}{
\begin{tabu} to \textwidth {X[c]|X[c]|X[c]|X[c]|X[c]|X[c]|X[c]|X[c]|X[c]|X[c]|X[c]}
\tabucline[1.5pt]{-}
\multicolumn{2}{c|}{Number of Frame Images} & \multicolumn{3}{c|}{TRAIN} & \multicolumn{3}{c|}{VAL} & \multicolumn{3}{c}{TEST} \\
\tabucline[1pt]{-}
\multicolumn{1}{c|}{State} & \multicolumn{1}{c|}{Total number} & \multicolumn{1}{c|}{Video 1} & \multicolumn{1}{c|}{Video 2} & \multicolumn{1}{c|}{Video 3} & \multicolumn{1}{c|}{Video 1} & \multicolumn{1}{c|}{Video 2} & \multicolumn{1}{c|}{Video 3} & \multicolumn{1}{c|}{Video 1} & \multicolumn{1}{c|}{Video 2} & \multicolumn{1}{c}{Video 3}  \\
\tabucline[1pt]{-}
initial & 36000 & 9000 & 9000 & 9000 & 1800 & 1800 & 1800 & 1200 & 1200 & 1200 \\
\tabucline[1pt]{-}
noise & 252000 & 63000 & 63000 & 63000 & 12600 & 12600 & 12600 & 8400 & 8400 & 8400 \\
\tabucline[1.5pt]{-}
\end{tabu}}}
\label{table_dataset}
\end{table*}

\noindent\textbf{Evaluation Metrics.}
Following previous methods~\cite{trades}, we employ CLEAR-MOT metrics~\cite{bernardin2008evaluating} to evaluate the tracking performance. This metric consists of various indicators, including MOTA, IDF1, IDs, FP, FN. The explanation of each indicator is introduced as follows:
\begin{enumerate}
    \item[-] MOTA is a comprehensive evaluation metric used to measure the overall performance of a tracker across the entire dataset;
    \item[-] IDF1 measures a tracker's ability to maintain correct identity labels throughout an entire sequence;
    \item[-] False Positives (FP) refer to instances where the tracking algorithm erroneously detects the presence of a target that does not exist in reality;
    \item[-] False Negatives (FN) refer to instances where the tracking algorithm fails to detect a target that is actually present;
    \item[-] Identity switches (IDs) occur when the tracking algorithm mistakenly changes the identifier of a target during the tracking process.
\end{enumerate}

Among these indicators, MOTA and IDF1 stand out as the most important for evaluating tracking performance.

\begin{table}[!t]
 \setlength{\tabcolsep}{9pt}
\normalsize
\centering
\renewcommand{\arraystretch}{1.0}
\caption{\small Performance of different algorithms on the no-noise phytoplankton dataset.}
\scalebox{0.7}{
\begin{tabu}{c|ccccc}
\tabucline[1.5pt]{-}
Method & IDF1$\uparrow$ & MOTA$\uparrow$ & FP$\downarrow$ & FN$\downarrow$ & IDs$\downarrow$
 \\
\tabucline[1.0pt]{-}
Sort (ICIP'16)\cite{sort} & 69.3 & 55.4 & \textbf{171} & 3829 & 392 \\
DeepSort (ICIP'17)\cite{deepsort} & 24.7 & 39.1 & 1906 & 4032 & 53 \\
DeepMot (CVPR'20)\cite{xu2019deepmot} & 16.9 & 40.8 & 2038 & 3737 & 52 \\
TraDeS (CVPR'21)\cite{trades} & 85.4 & 78.3 & 417 & 1641 & 81 \\
BotSort (Arxiv'22)\cite{aharon2022botsort} & 57.9 & 40.0 & 594 & 5239 & 70 \\
ByteTrack (ECCV'22)\cite{zhang2022bytetrack} & 84.6 & 71.8 & 588 & 2183 & \textbf{2} \\
UAVMot (CVPR'22)\cite{uavmot} & 83.2 & 69.7 & 601 & 2375 & 11 \\
StrongSort (TMM'23)\cite{du2023strongsort} & 48.3 & 37.0 & 792 & 4612 & 794 \\
UCMCTrack (AAAI'24)\cite{yi2024ucmc} & 61.2 & 56.4 & 153 & 2467 & 425 \\
BoostTrack (MVA'24)\cite{stanojevic2024boostTrack} & 86.4 & 75.6 & 306 & 1746 & 6 \\
TLTDMOT (CVPR'24)\cite{chen2024delving} & 85.2 & 74.3 & 307 & 1557 & 46 \\
\tabucline[1.0pt]{-}
\textbf{\textit{PhyTracker}} & \textbf{88.4} & \textbf{80.8} & 356 & \textbf{1482} & 55 \\
\tabucline[1.5pt]{-}
\end{tabu}}
\label{table_compare}
\end{table}

\smallskip
\noindent\textbf{Implementation Details.}
For training, the input image size was uniformly adjusted to 860 $\times$ 640, and the learning rate was decreased at the 40th and 50th epochs. We utilized the DLA34 network~\cite{yu2018deep} for feature extraction and trained on a single 4080 GPU for 60 epochs, with a batch size set to 5 and an initial learning rate of 2.5$\times$$10^{-4}$. In the inference phase, the tracking confidence threshold was set to 0.4, the input image size was set to 640 $\times$ 960, and input was based on the previous two consecutive frames simultaneously. In the comparative analysis of tracking-by-detection approaches, CenterNet is consistently utilized as the detection framework, with DLA34 serving as its backbone architecture. The remaining parts follow the default settings of TraDeS~\cite{trades}.

To enhance generalization capabilities and prevent overfitting, we incorporated data augmentation into our training process with a certain probability for each set of data. The primary techniques used include flipping, cropping, scaling, and translating, among others. Regarding training duration, both our method and the comparison method adhere to the same number of training epochs, which is 60. When training \textit{PhyTracker}, we standardized the input image size to 860 $\times$ 640 and reduced the learning rate at the 40th and 50th epochs.

We use the general objectives as in~\cite{trades}, which is defined as $\mathcal{L} = 
  \mathcal{L}_{\mathrm{det}} + \mathcal{L}_{\mathrm{mask}} + \mathcal{L}_{\mathrm{CVA}}$,
where $\mathcal{L}_{\mathrm{det}}$ is the detection loss~\cite{zhou2019objects}, $\mathcal{L}_{\mathrm{mask}}$ is the instance segmentation loss~\cite{tian2020conditional}, and a ReID loss~\cite{luo2019bag}.

\begin{table*}[!t]
\centering
\renewcommand{\arraystretch}{1.4}
\caption{\small Gray Processing is not applied in the medium difficulty level. In the hard difficulty level of Gray Processing, we convert all images to grayscale. In the table below, if the value of the MOTA metric is less than or equal to 0, it is represented by "-".}
\scalebox{1.0}{
\resizebox{\textwidth}{!}{
\begin{tabu}
{c@{\hspace{0.6em}}|c@{\hspace{0.6em}}|c@{\hspace{0.6em}}c@{\hspace{0.6em}}c@{\hspace{0.6em}}|c@{\hspace{0.6em}}c@{\hspace{0.6em}}c@{\hspace{0.6em}}c@{\hspace{0.6em}}}
\tabucline[1.5pt]{-}
\multicolumn{1}{c@{\hspace{0.6em}}}{MOTA} & \multicolumn{1}{|c@{\hspace{0.6em}}}{EASY} & \multicolumn{3}{|c@{\hspace{0.6em}}}{MEDIUM} & \multicolumn{4}{|c@{\hspace{0.6em}}}{HARD} \\
\cline{2-9}
\multicolumn{1}{c@{\hspace{0.6em}}}{} & \multicolumn{1}{|c@{\hspace{0.6em}}}{NONE} & \multicolumn{1}{|c@{\hspace{0.6em}}}{Occlusion} & \multicolumn{1}{c@{\hspace{0.6em}}}{Blurring} & \multicolumn{1}{c@{\hspace{0.6em}}}{Salt and Pepper} & \multicolumn{1}{|c@{\hspace{0.6em}}}{Occlusion} & \multicolumn{1}{c@{\hspace{0.6em}}}{Gray Processing} & \multicolumn{1}{c@{\hspace{0.6em}}}{Blurring} & \multicolumn{1}{c@{\hspace{0.6em}}}{Salt and Pepper}  \\
\tabucline[1.0pt]{-}
Sort       (ICIP'16)\cite{sort}               & 55.4          & 28.5          & 43.0          & 0.7          & 18.5          & 12.4          & 18.4          & 0.3          \\
DeepSort   (ICIP'17)\cite{deepsort}           & 39.1          &  -            & 24.1          &  -           &  -            &  -            & -          &  -           \\
DeepMot    (CVPR'20)\cite{xu2019deepmot}      & 40.8          & 2.8           & 32.3          & 0.2          &  -            &  -            & -          &  -           \\
TraDeS     (CVPR'21)\cite{trades}             & 78.3          & 52.0          & 44.1          & 0.7          & 45.1          & 27.5          & -          &  -           \\
BotSort    (Arxiv'22)\cite{aharon2022botsort}  & 40.0          & 16.4         & 9.5           & 0.3          & 9.7           & 9.5           & -          &  -           \\
ByteTrack  (ECCV'22)\cite{zhang2022bytetrack} & 71.8          & 36.2          & 56.8          & 1.1          & 24.4          & 17.1          & 12.1          & 0.5          \\
UAVMot     (CVPR'22)\cite{uavmot}             & 69.7          & 35.7          & 57.0          & 1.1          & 24.7          & 17.8          & 12.7          & 0.6          \\
StrongSort (TMM'23)\cite{du2023strongsort}   & 37.0          & 27.0           & 35.1          & 0.9          & 3.2           & 10.5          & -          & 0.2          \\
UCMCTrack  (AAAI'24)\cite{yi2024ucmc}         & 56.4          & 11.5          & 31.3          & 0.1          & 8.6           & 15.3          & -          &  -           \\
BoostTrack (MVA'24)\cite{stanojevic2024boostTrack} & 75.6    & 44.5          & \textbf{58.6}  &  1.3         & 36.4          & 28.1          & \textbf{26.1} & 0.6          \\
TLTDMOT    (CVPR'24)\cite{chen2024delving}    & 74.3          & 45.0          & 54.2          &  0.9         & 32.6          & 17.7          & 17.7          &  -           \\
\tabucline[1.0pt]{-}
\textbf{\textit{PhyTracker}}         & \textbf{80.8} & \textbf{52.4} & 53.2          & \textbf{1.4} & \textbf{45.4} & \textbf{31.9} & 24.9          & \textbf{0.8} \\
\tabucline[1.5pt]{-}
\end{tabu}}}
\label{table_full_compare}
\end{table*}

\begin{table}[!t]
 \setlength{\tabcolsep}{9pt}
\normalsize
\centering
\renewcommand{\arraystretch}{1.0}
\caption{\small In MOT17, \textit{PhyTracker} outperforms TraDeS in IDF1, MOTA, and FN metrics.}
\scalebox{0.8}{
\begin{tabu}{c|ccccc}
\tabucline[1.5pt]{-}
Method & IDF1$\uparrow$ & MOTA$\uparrow$ & FP$\downarrow$ & FN$\downarrow$ & IDs$\downarrow$
 \\
\tabucline[1.0pt]{-}
TraDeS \cite{trades} & 58.2 & 51.8 & \textbf{1091} & 2554 & \textbf{38} \\
\textbf{\textit{PhyTracker}} & \textbf{60.4} & \textbf{53.6} & 1123 & \textbf{2465} & 41 \\
\tabucline[1.5pt]{-}
\end{tabu}}
\label{table_mot}
\end{table}

\smallskip
\noindent\textbf{Compared Methods.}
To ensure the thoroughness of our experiments, we selected several methods with differing focuses for comparative analysis: 
SORT~\cite{sort}, DeepSORT~\cite{deepsort}, StrongSORT~\cite{du2023strongsort}, BotSORT~\cite{aharon2022botsort}, DeepMOT~\cite{xu2019deepmot}, ByteTrack~\cite{zhang2022bytetrack}, and UAVMOT~\cite{uavmot}. The introduction of each method is as follows:
\begin{enumerate}
    \item[-] SORT (ICIP'16)~\cite{sort}: SORT is a simple and real-time multi-object tracking algorithm that uses a Kalman filter and the Hungarian algorithm for data association, significantly enhancing tracking accuracy and speed.
    \item[-] DeepSORT (ICIP'17)~\cite{deepsort}: DeepSORT is an algorithm that builds upon SORT by incorporating ReID (Re-identification) and appearance features of detection boxes. It utilizes a Matching Cascade approach to reduce the number of target ID switches.
    \item[-] DeepMOT (CVPR'20)~\cite{xu2019deepmot}: DeepMOT introduces a deep Hungarian network that uses dual bidirectional recurrent neural networks (BRNN)~\cite{schuster1997bidirectional} to convey global information within the cost matrix. Its loss function is designed based on two differentiable evaluation metrics, which optimize the network's output allocation matrix to enhance accuracy.
    \item[-] TraDeS (CVPR'21)~\cite{trades}: TraDeS is an online joint detection and tracking model that utilizes tracking cues to aid in end-to-end detection. It infers target tracking offsets from cost measures, which are used to propagate previous target features to improve current target detection and segmentation.
    \item[-] BotSORT (Arxiv'22)~\cite{aharon2022botsort}: BotSORT integrates the advantages of motion and appearance information with camera motion compensation and a more precise Kalman filter state vector, achieving accurate tracking results. 
    \item[-] ByteTrack (ECCV'22)~\cite{zhang2022bytetrack}: ByteTrack employs a multi-match approach during tracking. It initially matches high-scoring detection boxes with existing tracks, then matches lower-scoring boxes with tracks that were not matched in the first round, addressing the scenario of object occlusions. Additionally, it solely uses the Kalman filter and Hungarian algorithm, eliminating the need for a ReID model.
    \item[-] UAVMOT (CVPR'22)~\cite{uavmot}: UAVMOT is an algorithm for drone viewpoints, based on improvements over FairMOT. It incorporates three key components: an ID feature update module, an adaptive motion filter, and gradient balanced focal loss. These enhancements respectively strengthen the reID feature connections between adjacent frames, address the complex motion in drone video footage, and optimize the training of heatmaps.
    \item[-] StrongSORT (TMM'23)~\cite{du2023strongsort}: StrongSORT has made a series of improvements over DeepSORT, such as improving the appearance feature extractor, introducing an inertia term to smooth feature updates, utilizing a Kalman filter designed for nonlinear motion, and adding a cost matrix that incorporates motion information.
    \item[-] UCMCTrack (AAAI'24)~\cite{yi2024ucmc}: In multi-object tracking, irregular camera motion has always been a challenge. This is because the rapid movement of the camera can cause abrupt changes in the positions of the objects in the frame, making it difficult to associate them with their past trajectories. UCMCTrack establishes a connection between the motion state of the objects and the ground in the image. It employs a mapped Mahalanobis Distance as an alternative to IoU to measure the similarity between the objects and their trajectories.
    \item[-] BoostTrack (MVA'24)~\cite{stanojevic2024boostTrack}: Handling unreliable detections and avoiding identity switches are crucial for the success of multi-object tracking. BoostTrack incorporates a confidence score for detection tracklets and utilizes it to scale the similarity measure. To reduce ambiguities caused by using IoU, BoostTrack proposes a novel addition of Mahalanobis distance and shape similarity to enhance the overall similarity measure.
    \item[-] TLTDMOT (CVPR'24)~\cite{chen2024delving}: TLTDMOT introduces a series of strategies to address the long-tail distribution problem in the field of multi-object tracking. These strategies include two data augmentation techniques, namely Static Camera Viewpoint Augmentation and Dynamic Camera Viewpoint Augmentation. Additionally, TLTDMOT incorporates a Group Softmax module for re-identification purposes.
\end{enumerate}

These methods include both tracking by detection and end-to-end tracking approaches. Their improvement directions vary, with some focusing on filtering detection boxes and some on addressing camera shake issues. 
All these methods are trained and tested under a same configuration with our method.

\begin{figure*}[!t]
  \centering
  \includegraphics[width=1.0\linewidth]{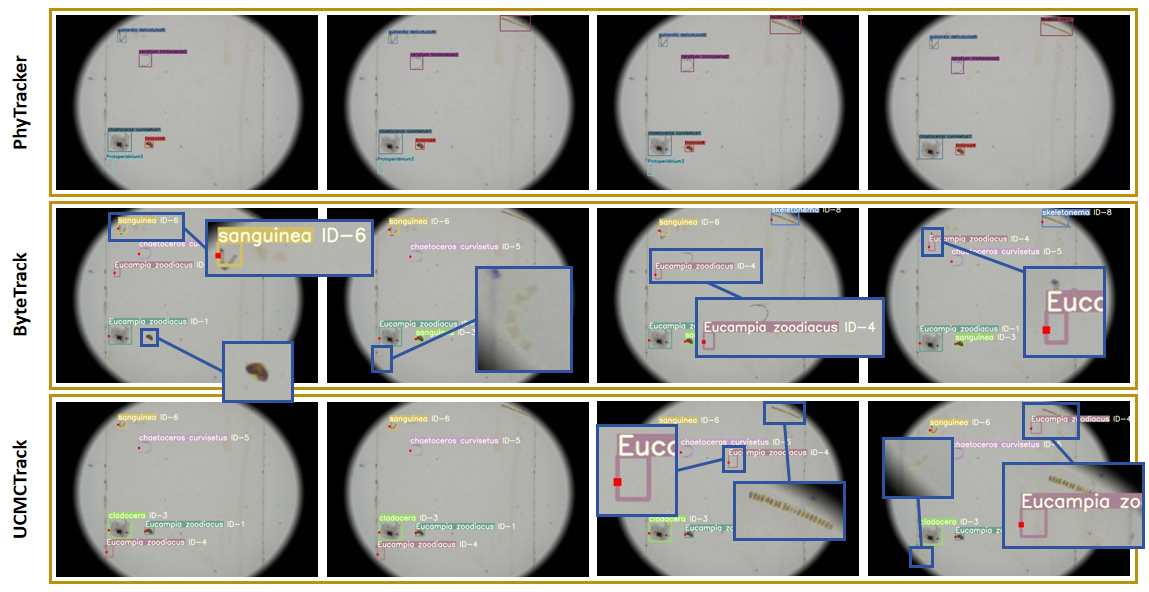}
  \caption{\small Each row in the figure represents four consecutive frames. The first row is \textit{PhyTracker}, the second row is ByteTrack, and the third row is UCMCTrack. The category names from top to bottom are: sanguinea, chaetoceros curvisetus, cladocera, sanguinea, eucampia zoodiacus.}
  \label{fig:compare1}
\end{figure*}

\subsection{Experimental Analysis}
We evaluated 12 different methods on the phytoplankton dataset and the MOT17 dataset.
For detailed metrics of the testing performance of the twelve methods on the phytoplankton dataset without noise, refer to Table~\ref{table_compare}. MOTA and IDF1 metrics reflect the overall tracking performance, while FN and FP indicate detection performance. Among the five evaluation metrics, \textit{PhyTracker} achieved the best results in IDF1, MOTA, and FN. Our ratings in FP and IDs are also better than most algorithms. The overall comparison results in the phytoplankton dataset are illustrated in Table~\ref{table_full_compare}. \textit{PhyTracker} achieved the best results in the dataset under noise-free conditions, as well as in scenarios with added noise, including occlusion, gray processing and salt-and-pepper noise. 
In the blurring data, \textit{PhyTracker} performs slightly worse than BoostTrack~\cite{stanojevic2024boostTrack}. This is because the blurring process degrades the image quality, leading to poorer feature extraction by the network. End-to-end algorithms like \textit{PhyTracker} directly track using the extracted features, making them more susceptible to the effects of blurring. In contrast, tracking-by-detection algorithms like BoostTrack first generate bounding boxes and then pass them to the tracking component for further processing. During this process, bounding boxes with low confidence are not completely discarded, which reduces the impact of blurring on tracking-by-detection algorithms.
In contrast to earlier methods such as DeepSort~\cite{deepsort}, our approach employs an end-to-end model, eliminating reliance on detection accuracy. Unlike recent methods like TLTDMOT~\cite{chen2024delving}, \textit{PhyTracker} focuses more on the unique characteristics of phytoplankton data. By enhancing the feature representation and association of phytoplankton while mitigating the impact of moving features on overall recognition, \textit{PhyTracker} consistently outperforms other methods.

\begin{figure*}[!t]
  \centering
  \includegraphics[width=1.0\linewidth]{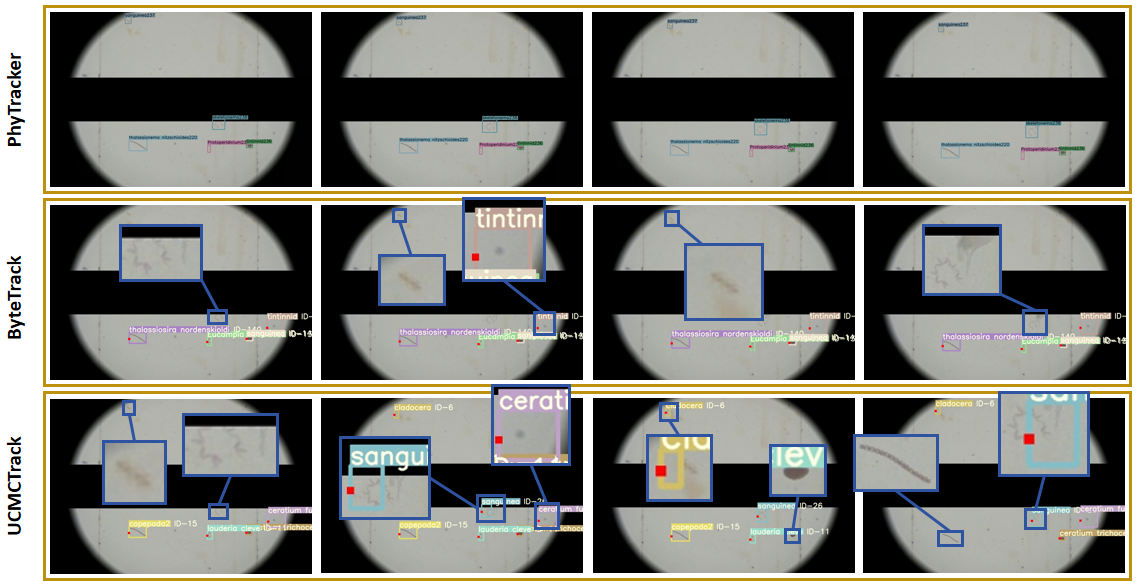}
  \caption{\small Each row in the figure shows four consecutive frames: the first row is \textit{PhyTracker}, the second row is ByteTrack, and the third row is UCMCTrack. The category names from top to bottom are: sanguinea, skeletonema, thalassionema nitzschioides, protoperidinium, tintinnid.}
  \label{fig:compare2}
\end{figure*}

To represent the differences between \textit{PhyTracker} and other methods, we tested the data without added noise using \textit{PhyTracker}, ByteTrack~\cite{zhang2022bytetrack} and UCMCTrack~\cite{yi2024ucmc}, with the results shown in Fig.~\ref{fig:compare1}. As evident from the comparison, mainstream multi-object tracking methods do not perform well on phytoplankton data. The tracking performance of \textit{PhyTracker}, ByteTrack, and UCMCTrack under occlusion conditions is shown in Fig.~\ref{fig:compare2}. After adding occlusion, the performance of ByteTrack and UCMCTrack decreased, but \textit{PhyTracker} still performed well. Compared to these two methods, \textit{PhyTracker}, which relies on the FMR module capable of associating all previous frames, is better suited for tracking plankton under more challenging conditions. 

We conducted a comparison between TraDeS~\cite{trades} and \textit{PhyTracker} in MOT17 dataset, with the results illustrated in the provided Table~\ref{table_mot}. Training under the same conditions, \textit{PhyTracker} performs better than TraDeS. This indicates that our improvements based on the characteristics of phytoplankton are still effective in pedestrian data, especially regarding the strategies for feature enhancement and strengthened association for tracking objects.

\begin{table}[htbp]
 \setlength{\tabcolsep}{9pt}
\normalsize
\centering
\renewcommand{\arraystretch}{1.0}
\caption{\small We tested the effectiveness of the three modules under different embedding scenarios, validating the efficacy of each module.}
\scalebox{0.8}{
\begin{tabu}{l|ccc|cc}
\tabucline[1.5pt]{-}
Scheme & TFE & ATA & FMR & IDF1$\uparrow$ & MOTA$\uparrow$
 \\
\tabucline[1.0pt]{-}
1 Baseline &  &  &  & 85.4 & 78.3 \\
2 & $\checkmark$ &  &  & 87.2 & 80.2 \\
3 & $\checkmark$ & $\checkmark$ &  & 87.2 & 80.4 \\
4 & $\checkmark$ &  & $\checkmark$ & 87.5 & 80.5 \\
5 &  & $\checkmark$ & $\checkmark$ & 86.7 & 79.5 \\
6 \textbf{\textit{PhyTracker}} & $\checkmark$ & $\checkmark$ & $\checkmark$ & \textbf{88.4} & \textbf{80.8} \\
\tabucline[1.5pt]{-}
\end{tabu}}
\label{table_full_module}
\end{table}

\subsection{Ablation Study}
In this section, we validate the effectiveness of \textit{PhyTracker} through ablation studies. All experiments were conducted on the phytoplankton dataset and its noise-augmented sections. To ensure a fair evaluation of each component's performance, our training and testing details are consistent with those described in the Implementation Details section above. When testing a specific module, no changes were made to the remaining modules. 
Compared to TraDeS, \textit{PhyTracker} focuses more on the feature representation, enhanced association, and trajectory influence of plankton. We implemented these aspects respectively in TFE, ATA, and FMR.
To verify the effectiveness of the three modules we proposed, we conducted five sets of experiments. These involved incorporating only TFE under the same conditions, adding both TFE and ATA, adding both TFE and FMR, adding both ATA and FMR, and finally integrating all three modules. As shown in the Table~\ref{table_full_module}, there is a noticeable increase in scores after incorporating each module, compared to the baseline alone. With all three modules integrated, in comparison to the baseline, MOTA and IDF1 scores increased by 2.5\% and 3\%, respectively.

\smallskip
\noindent\textbf{Effect of TFE.}
We tested the effectiveness of the TFE module details using phytoplankton data and conducted three sets of experiments for comparison: 

\textit{1) Study on SRM Filters.} In the TFE module, we employed three fixed SRM convolutional kernels for detecting weak edges, strong edges, and sharpening, respectively. As shown in Eq.~\ref{eq: srm}. To validate the effectiveness of this combination, we compared it with another set of three fixed commonly used SRM convolutional kernels: horizontal edge detection, vertical edge detection, and sharpening. The results are shown in Table~\ref{table_srm}. Compared to commonly used SRM convolution kernels, the improved SRM convolution kernel we use has improved IDF1 and MOTA metrics by 1.9\% and 1.2\%, respectively. The commonly used convolution kernels for comparison are as follows: 
\begin{equation}
{
\setlength{\arraycolsep}{1.2pt}
\begin{aligned}
\small
\begin{bmatrix}
  -1 &  -2 &  -4 &  -2 &  -1 \\
  0 &  0 &  0 &  0 &  0 \\
  0 &  0 &  0 &  0 &  0 \\
  0 &  0 &  0 &  0 &  0 \\
  1 &  2 &  4 &  2 &  1 \\
\end{bmatrix}
\begin{bmatrix}
  -1 &  0 & 0 &  0 &  1 \\
  -2 &  0 & 0 &  0 &  2 \\
  -4 &  0 & 0 &  0 &  4 \\
  -2 &  0 & 0 &  0 &  2 \\
  -1 &  0 & 0 &  0 &  1 \\
\end{bmatrix}
\begin{bmatrix}
  0 &  0 & -1 &  0 &  0 \\
  0 &  0 & -1 &  0 &  0 \\
 -1 & -1 &  9 & -1 & -1 \\
  0 &  0 & -1 &  0 &  0 \\
  0 &  0 & -1 &  0 &  0 \\
\end{bmatrix}
\end{aligned}
}
\end{equation}

\begin{table}[htbp]
 \setlength{\tabcolsep}{9pt}
\normalsize
\centering
\renewcommand{\arraystretch}{1.0}
\caption{\small Mode1 is ours, and Mode2 is another set of convolutional kernels used for comparison.}
\scalebox{0.8}{
\begin{tabu}{c|cc|cc}
\tabucline[1.5pt]{-}
\textit{PhyTracker} & Mode1 & Mode2 & IDF1$\uparrow$ & MOTA$\uparrow$
 \\
\tabucline[1.0pt]{-}
SRM &  & $\checkmark$ & 86.5 & 79.6 \\
SRM & $\checkmark$ &  & \textbf{88.4} & \textbf{80.8} \\
\tabucline[1.5pt]{-}
\end{tabu}}
\label{table_srm}
\end{table}

\textit{2) Study on TFE module.} To validate the significant effectiveness of the TFE module on phytoplankton data, we conducted a comparison using the CBAM module, which integrates channel and spatial attention mechanisms. In this experiment, we replaced the TFE module in the model with the CBAM module, keeping all other parts unchanged. Using the TFE module compared to the CBAM module improved IDF1 and MOTA by 3.1\% and 2.7\%, respectively. As shown in the Table~\ref{table_tfe}. Compared to other commonly used modules, the TFE module has advantages in extracting plankton features due to its unique structure and specifically designed SRM. 
\begin{table}[htbp]
 \setlength{\tabcolsep}{9pt}
\normalsize
\centering
\renewcommand{\arraystretch}{1.0}
\caption{\small The first line indicates the use of CBAM, and the second line indicates the use of TFE.}
\scalebox{0.8}{
\begin{tabu}{c|cc|cc}
\tabucline[1.5pt]{-}
Baseline+ATA+FMR & CBAM & TFE & IDF1$\uparrow$ & MOTA$\uparrow$
 \\
\tabucline[1.0pt]{-}
+ & $\checkmark$ &  & 85.3 & 78.1 \\
+ &  & $\checkmark$ & \textbf{88.4} & \textbf{80.8} \\
\tabucline[1.5pt]{-}
\end{tabu}}
\label{table_tfe}
\end{table}

\textit{3) Study on SIE module.} In the TFE module, we utilize the SIE module to extract semantic information and enhance the feature representation of tracking targets. To validate the effectiveness of the SIE module, we removed it from the TFE module and conducted training under the same conditions. The results, as shown in the Table~\ref{table_sie}, indicate that after removing the SIE module, the IDF1 and MOTA scores dropped by 1.4\% and 0.9\%, respectively. 
Compared to using backbone network alone for image feature extraction in TraDeS, the SIE module first employs 3$\times$3 convolutions and SRM convolutions to extract semantic information specific to phytoplankton. This enriched information is then passed into the backbone network for further feature extraction, resulting in more comprehensive corresponding features.

\begin{table}[htbp]
 \setlength{\tabcolsep}{9pt}
\normalsize
\centering
\renewcommand{\arraystretch}{1.0}
\caption{\small We removed the SIE sub-module from the TFE module while keeping all other conditions unchanged to verify the effect of SIE.}
\scalebox{0.8}{
\begin{tabu}{c|c|cc}
\tabucline[1.5pt]{-}
\textit{PhyTracker} & SIE & IDF1$\uparrow$ & MOTA$\uparrow$
 \\
\tabucline[1.0pt]{-}
TFE & $\times$ & 87.0 & 79.9 \\
TFE & $\checkmark$ & \textbf{88.4} & \textbf{80.8} \\
\tabucline[1.5pt]{-}
\end{tabu}}
\label{table_sie}
\end{table}

\smallskip
\noindent\textbf{Effect of ATA.}
In the ATA module, we refine the past frame feature $f^{t-1}$ to compute query, and refine the current frame feature $f^t$ to compute key and value. Attention computation is then performed in the second layer. To verify the effectiveness of this structure, we conducted two sets of experiments. \textbf{1)} The overall structure is modified to sequentially perform attention computation twice, with the input to the second layer being the single tensor output from the first layer. \textbf{2)} We exchanged the query, key, and value calculated for the past and current frames. As shown in Table~\ref{table_ata}, after altering the structure, the IDF1 and MOTA scores dropped by 1.1\% and 0.4\%, respectively. After swapping the variables corresponding to the two frames, the IDF1 and MOTA scores dropped by 2.0\% and 0.7\%, respectively. 
This indicates that simply applying or stacking transformer does not effectively suit the application of phytoplankton data and can cause the algorithm's focus to shift. To eliminate this bias, we first refine the features of each frame and then apply attention calculations only to the refined features, thereby minimizing the influence of non-tracking targets as much as possible. 

\begin{table}[htbp]
 \setlength{\tabcolsep}{9pt}
\normalsize
\centering
\renewcommand{\arraystretch}{1.0}
\caption{\small Exp1 corresponds to \textbf{1)}, exp2 corresponds to \textbf{2)}, and exp3 is our method.}
\scalebox{0.8}{
\begin{tabu}{c|ccc|cc}
\tabucline[1.5pt]{-}
\textit{PhyTracker} & Exp1 & Exp2 & Exp3 & IDF1$\uparrow$ & MOTA$\uparrow$
 \\
\tabucline[1.0pt]{-}
ATA & $\checkmark$ &  &  & 87.3 & 80.4 \\
ATA &  & $\checkmark$ &  & 86.4 & 80.1 \\
ATA &  &  & $\checkmark$ & \textbf{88.4} & \textbf{80.8} \\
\tabucline[1.5pt]{-}
\end{tabu}}
\label{table_ata}
\end{table}

\smallskip
\noindent\textbf{Effect of FMR.}
To verify the necessity of the dual-branch structure in the FMR module and the rationality of its computational design, we conducted two sets of experiments. \textbf{1)} We removed the tracking offset branch for the current frame, retaining only the branch used to eliminate past motion characteristics. \textbf{2)} We modified the matrix calculation for removing past characteristics, specifically changing the calculation of memory offset to $\lambda_t = {\mathcal{O}_t} + \lambda_{t-1}$. Removing one branch resulted in IDF1 and MOTA scores dropping by 1.0\% and 0.5\%, respectively. After modifying the matrix calculation, IDF1 and MOTA scores dropped by 0.4\% and 0.3\%, respectively. As shown in Table~\ref{table_fmr}, When we remove the tracking offset of the current frame, the tracker's performance significantly declines due to the lack of this feature. However, unreasonable feature fusion can also negatively impact the tracker. Our method has achieved optimal results overall.

\begin{table}[htbp]
 \setlength{\tabcolsep}{9pt}
\normalsize
\centering
\renewcommand{\arraystretch}{1.0}
\caption{\small Exp1 corresponds to \textbf{1)}, exp2 corresponds to \textbf{2)}, and exp3 is our method.}
\scalebox{0.8}{
\begin{tabu}{c|ccc|cc}
\tabucline[1.5pt]{-}
\textit{PhyTracker} & Exp1 & Exp2 & Exp3 & IDF1$\uparrow$ & MOTA$\uparrow$
 \\
\tabucline[1.0pt]{-}
FMR & $\checkmark$ &  &  & 87.4 & 80.3 \\
FMR &  & $\checkmark$ &  & 88.0 & 80.5 \\
FMR &  &  & $\checkmark$ & \textbf{88.4} & \textbf{80.8} \\
\tabucline[1.5pt]{-}
\end{tabu}}
\label{table_fmr}
\end{table}

\section{Conclusion}
In this paper, we develop PyTracker, an intelligent in situ tracking framework to address the critical challenges of phytoplankton monitoring. Unlike traditional non-in situ methods, PyTracker offers an automated and timely solution for monitoring phytoplankton, thereby enhancing the efficiency and accuracy of ecosystem analysis. Confronting difficulties in this task, including their constrained mobility, inconspicuous appearance, and the presence of impurities in water samples, our method proposes three novel modules: the Texture-enhanced Feature Extraction (TFE) module for improved feature capture, the Attention-enhanced Temporal Association (ATA) module for distinguishing phytoplankton from impurities, and the Flow-agnostic Movement Refinement (FMR) module for refining movement characteristics. These innovations collectively enhance the tracking performance and reliability of the system.

Through extensive experiments on the PMOT dataset, PyTracker has demonstrated superior performance in tracking phytoplankton. Moreover, our method has shown its versatility and effectiveness on the MOT dataset, surpassing conventional tracking methods. This work not only underscores the importance of tailored tracking solutions for aquatic environments but also sets a foundation for future advancements in marine ecological monitoring and scientific exploration.

\section{Acknowledgments}
This work was supported in part by the National Key R\&D Program of China under Grant 2022ZD0117201.

\small
\bibliographystyle{unsrt}
\bibliography{reference}

\begin{IEEEbiography}[{\includegraphics[width=1in,height=1.25in,clip,keepaspectratio]{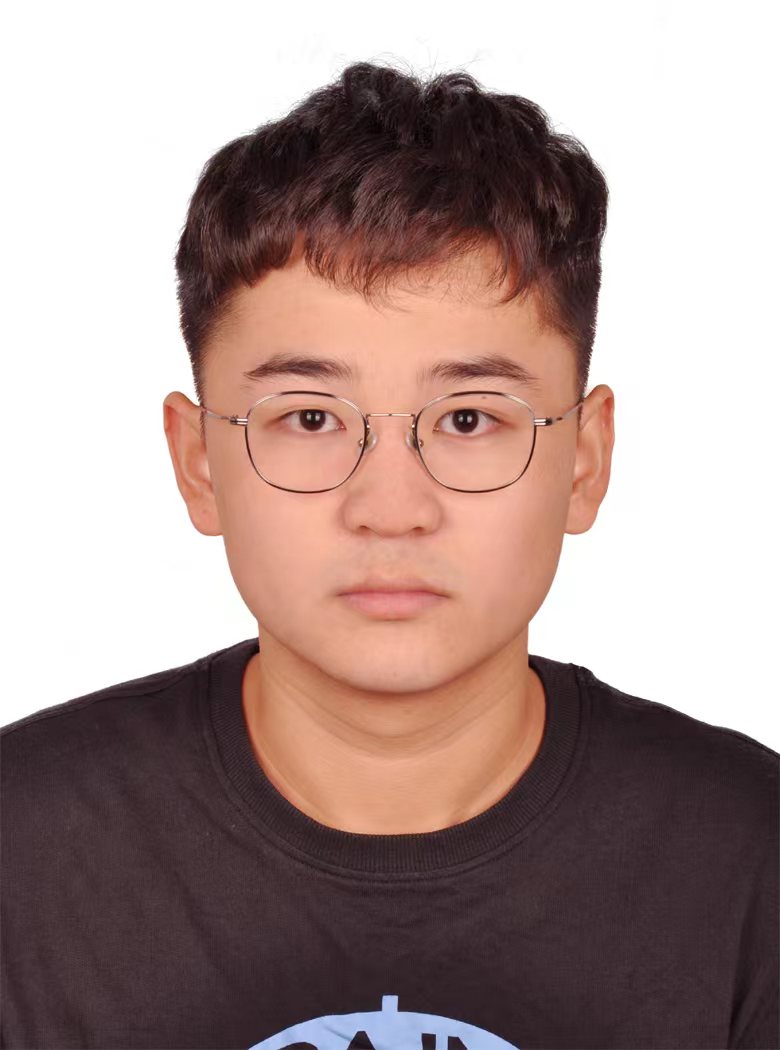}}]{Yang Yu}
	received the B.S. degree from Qufu Normal University, Qufu, China, in 2021, where he is currently pursuing the M.S. degree with the School of Computer Science and Technology, Ocean University of China, China. His current research interests include deep learning and multiple object tracking.
	
\end{IEEEbiography}

\begin{IEEEbiography}[{\includegraphics[width=1in,height=1.25in,clip,keepaspectratio]{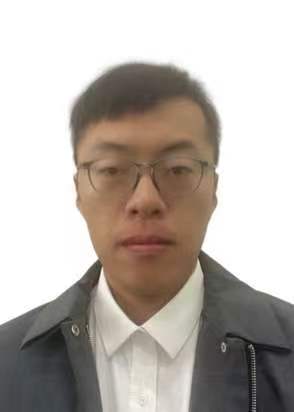}}]{Qingxuan Lv}
    was born in Shanxi, China, in 1996. He received his bachelor's degree in Computer Science and Technology from the Shanxi University of Finance and Economics in 2018. He received his master's degree in Computer Science and Technology from the Ocean University of China (OUC) in 2021. He is currently a candidate of a doctor's degree at the ocean group of VisionLab OUC. His research interests include computer vision and machine learning. Specifically, he is interested in universal domain adaptation and semantic segmentation.
	
\end{IEEEbiography}

\begin{IEEEbiography}[{\includegraphics[width=1in,height=1.25in,clip,keepaspectratio]{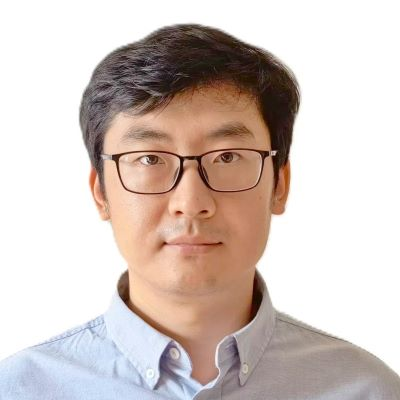}}]{Yuezun Li}
	(Member, IEEE) received the B.S. degree in Software Engineering from Shandong University in 2012, the M.S. degree in Computer Science in 2015, and the Ph.D. degree in computer science from University at Albany–SUNY, in 2020. He was a Senior Research Scientist with the Department of Computer Science and Engineering, University at Buffalo–SUNY. He is currently a Lecturer with the Center on Artifcial Intelligence, Ocean University of China. His research interests include computer vision and multimedia forensics. His work has been published in peer reviewed conferences and journals, including NeurIPS, ICCV, CVPR, TIFS, TCSVT, etc. 

\end{IEEEbiography}

\begin{IEEEbiography}[{\includegraphics[width=1in,height=1.25in,clip,keepaspectratio]{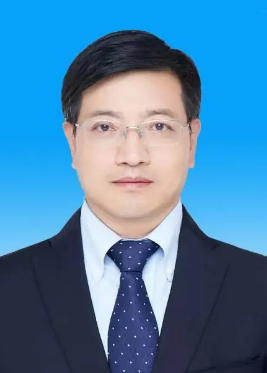}}]{Zhiqiang Wei}
	received the B.S. degree from Shandong University of Technology in 1992, the M.S. degree from Harbin Institute of Technology in 1995, and the Ph.D. degree from Tsinghua University in 2001. He was a professor and doctoral supervisor at Ocean University of China, as well as a visiting scholar funded by the China Scholarship Council at Carnegie Mellon University in the United States and the University of Manchester in the United Kingdom. He is currently the president of Qingdao University. His research interests include intelligent robotics technology, intelligent information processing, the Internet of Things and cloud computing, and multimedia big data technology.
	
\end{IEEEbiography}

\begin{IEEEbiography}[{\includegraphics[width=1in,height=1.25in,clip,keepaspectratio]{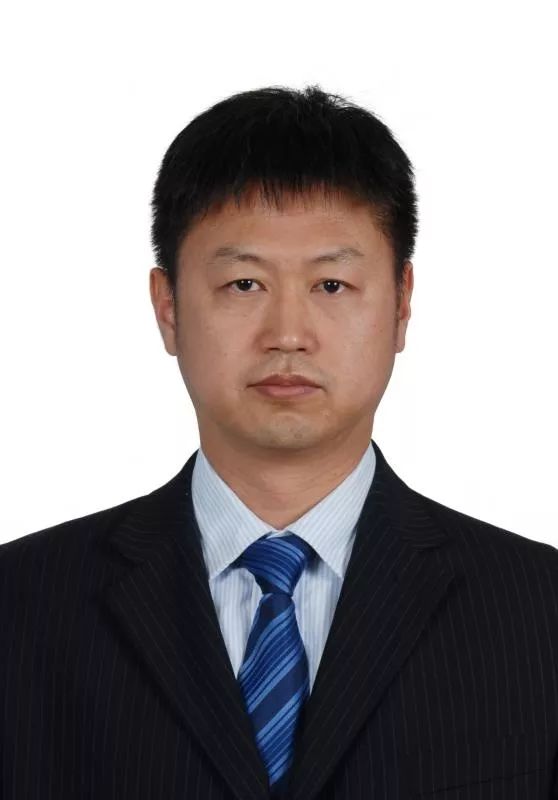}}]{Junyu Dong}
	received the B.Sc. and M.Sc. degrees in applied mathematics from the Department of Applied Mathematics, Ocean University of China, Qingdao, China, in 1993 and 1999, respectively, and the Ph.D. degree in image processing from the Department of Computer Science, Heriot-Watt University, Edinburgh, U.K., in November 2003. He is currently a Professor and the Minister of Faculty of Information Science and Engineering, Ocean University of China. His research interests include machine learning, big data, computer vision, and underwater image processing.
	
\end{IEEEbiography}

\vfill

\end{document}